\documentclass[sigconf]{acmart}
\acmConference[MM '23] {Proceedings of the 31st ACM International Conference on Multimedia}{October 29--November 3, 2023}{Ottawa, ON, Canada.}

\definecolor{Strawberry}{rgb}{1,0.26,0.64}
\definecolor{OliveGreen}{rgb}{0.0,0.69,0.31}
\definecolor{HaveLockBlue}{rgb}{0.26, 0.45, 0.76}

\usepackage{xspace}
\newcommand{\bodymethodname}{\textcolor{black}{GraMMaR}\xspace}
\newcommand{\bodyfullmethodname}{{\textbf{Gr}ound-\textbf{a}ware \textbf{M}otion \textbf{M}odel for 3D Hum\textbf{a}n Motion \textbf{R}econstruction}\xspace}

\usepackage{multirow}
\usepackage{xcolor}
\usepackage{balance}

\begin{document}
\title{
GraMMaR: Ground-aware Motion Model for 3D Human Motion Reconstruction
}

\author{Sihan Ma}
\email{sima7436@uni.sydney.edu.au}
\affiliation{
  \institution{The University of Sydney}
  \city{Sydney}
  \country{Australia}
}

\author{Qiong Cao}
\email{mathqiong2012@gmail.com}
\affiliation{
  \institution{JD Explore Academy}
  \city{Beijing}
  \country{China}
}

\author{Hongwei Yi}
\email{hongwei.yi@tuebingen.mpg.de}
\affiliation{
  \institution{Max Planck Institute for Intelligent Systems}
  \city{Tübingen}
  \country{Germany}
}

\author{Jing Zhang}
\email{jing.zhang1@sydney.edu.au}
\affiliation{
  \institution{The University of Sydney}
  \city{Sydney}
  \country{Australia}
}

\author{Dacheng Tao}
\email{dacheng.tao@gmail.com}
\affiliation{
  \institution{The University of Sydney}
  \city{Sydney}
  \country{Australia}
}

\begin{abstract}
Demystifying complex human-ground interactions is essential for accurate and realistic 3D human motion reconstruction from RGB videos, as it ensures consistency between the humans and the ground plane.
Prior methods have modeled human-ground interactions either implicitly or in a sparse manner, often resulting in unrealistic and incorrect motions when faced with noise and uncertainty. In contrast, our approach explicitly represents these interactions in a dense and continuous manner.
To this end, we propose a novel \bodyfullmethodname, named \textbf{\bodymethodname}, which jointly learns the distribution of transitions in both pose and interaction between every joint and ground plane at each time step of a motion sequence. 
It is trained to explicitly promote consistency between the motion and distance change towards the ground. 
After training, we establish a joint optimization strategy that utilizes \bodymethodname as a dual-prior, regularizing the optimization towards the space of plausible ground-aware motions. This leads to realistic and coherent motion reconstruction, irrespective of the assumed or learned ground plane.
Through extensive evaluation on the AMASS and AIST++ datasets, our model demonstrates good generalization and discriminating abilities in challenging cases including complex and ambiguous human-ground interactions. 
The code will be available at https://github.com/xymsh/GraMMaR.
\end{abstract}


\begin{teaserfigure}
    \centering
    \includegraphics[width=\textwidth]{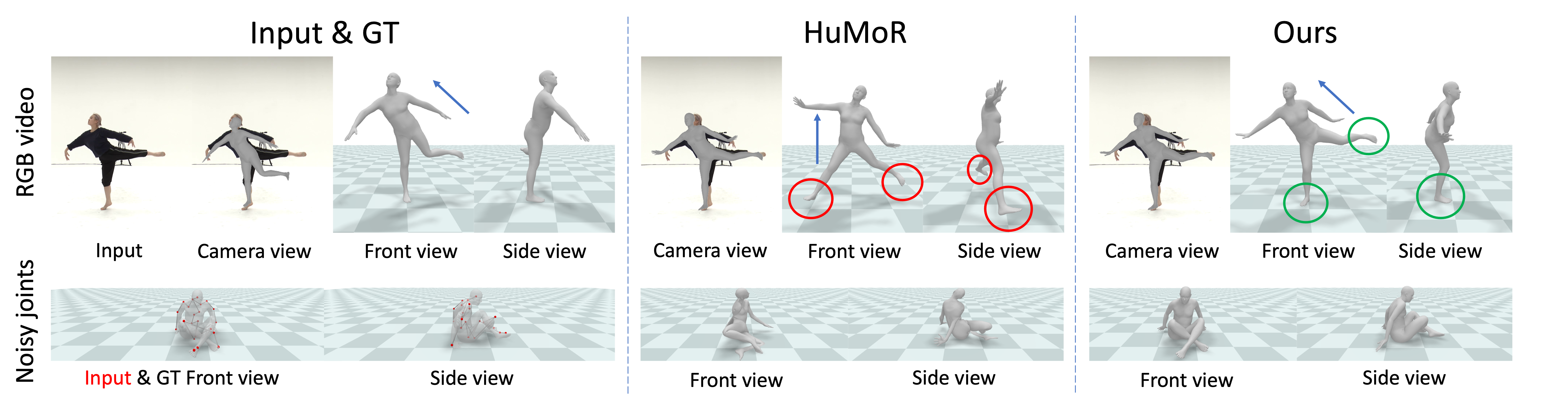}
    \caption{
    3D motion in the camera view is misleading. A representative optimization method HuMoR \cite{humor} produces correct poses under camera view but physically implausible poses in world view when faced with ambiguity (Row1) and noise (Row2). In contrast, our method provides a ground-aware motion, thereby ensuring physical plausibility across all views. 
    \textcolor{HaveLockBlue}{Body torso direction} and contacts for \textcolor{red}{HuMoR} and \textcolor{OliveGreen}{ours} are highlighted. GT in Row1 is reconstructed from multi-view images.
    }
    \label{fig:teaser}
\end{teaserfigure}

\maketitle

\section{Introduction}
The human body frequently engages in movements that involve interactions with the ground plane. In real-life scenarios, when a body part is in close proximity to the ground, individuals may need to slow down, lean their torso, orient their head to look at the ground or position their hands and feet on the ground. The capability to accurately predict 3D human motion with physical plausibility from RGB videos, which encompasses realistic interactions with the ground, is crucial for numerous applications~\cite{zhang2020empowering}, such as scene understanding~\cite{scene_understanding1,scene_understanding2,jperceiver}, 3D dance motion reconstruction and generation~\cite{dance_motion1,dance_motion2,dance_motion3,dance_motion4}, and augmented and virtual reality games~\cite{arvr1, arvr2, arvr3,terra}. While extensive research has focused on 3D motion estimation under camera space\cite{smplify, kevin-physics}, considering alignment solely in the camera view might be insufficient and potentially deceptive. There are cases where poses appear reasonable in the camera view but exhibit physically implausible body support on the assumed ground plane when viewed from an alternate viewpoint or placed in a 3D scene. Moreover, even within the camera view, handling noisy observations can result in visually implausible recovered motions, such as body twists, penetration, and jittery movements. Fig.~\ref{fig:teaser} demonstrates these issues. These challenges primarily arise because most state-of-the-art methods rarely consider the interaction between humans and the ground plane, thus unable to satisfy the physical constraints that govern the human body during interactions.

To address these issues, a natural solution is to model human-ground interaction explicitly to ensure consistency between the human body and the ground. Essentially, human-ground interaction involves the interdependent relationships between a 3D human and the ground plane. However, to date, few methods have explored human-ground interaction; those that do have primarily focused on the body-ground contact using binary contact labels \cite{humor, contact-dynamics} or ground reaction force \cite{kevin-physics}. \cite{contact-dynamics} models the interaction by directly predicting binary contact labels, indicating whether predefined joints are in contact with the ground. These classification results are then used as hard constraints that restrict the distance between joints and the ground during inference, thereby enabling the generation of physically plausible poses. However, the use of binary labels is inadequate as it only applies to joints in direct contact with the ground, leaving most other joints without physical restrictions. Moreover, the sparse and uncertain nature of contact occurrence across all joints significantly impacts the accuracy of motion reconstruction. The performance heavily relies on the quantity of frames and joints within a given motion sequence where contact is established, leading to instability. Alternatively, some work \cite{kevin-physics} has introduced ground reaction force as a means of representing human-ground interaction. A larger reaction force corresponds to a heavier penalty on the distance between the joints and the ground during optimization. Although intuitive, it is difficult to access and only applies to joints in contact with the ground.

In this work, we address these issues by building a robust human motion model that accurately captures the dynamics of 3D human motion through human-ground interactions. To achieve this goal, we first introduce a novel continuous distance-based per-joint interaction representation to encode fine-grained human-ground interactions at each time step. It overcomes the limitations of binary contact labels and ground reaction force by combining per-joint ground distance and its velocity along the gravity axis. Unlike previous methods, our new representation provides a continuous and differentiable measure with physical significance, allowing for a comprehensive depiction of motion patterns and ground-based body support for both contacting and non-contacting joints.

Building upon the novel representation, we devise an explicit ground-aware motion dynamics model that incorporates human-ground interactions and human pose. This is formulated as an autoregressive conditional variational auto-encoder (CVAE) \cite{cvae} to capture the temporal variations in human pose and human-ground interactions. The model simultaneously learns the distribution of transitions for both pose and joint-to-ground distances across adjacent frames within a motion sequence, producing a wide range of plausible poses and human-ground interactions. By conditioning the decoder to predict future motion based on existing poses and human-ground interactions, the model enforces consistency between the body and the ground plane.

We train our model on AMASS \cite{amass} and develop a joint optimization strategy for 3D human motion reconstruction from noisy observations and RGB videos. The trained model serves as a dual-prior to regularize the optimization towards the space of plausible ground-aware motions, resulting in realistic and coherent motion reconstruction, regardless of the assumed or learned ground plane. The resultant reconstruction method is termed \bodymethodname, which stands for \bodyfullmethodname. We evaluate \bodymethodname quantitatively and qualitatively on both RGB videos and noisy settings and demonstrate its superiority over the baseline in complex and ambiguous contact conditions. \bodymethodname proves effective irrespective of the ground plane being known or unknown.

\section{Related Work}

\textbf{Kinematic estimation.}
Kinematic methods for 3D pose estimation in videos~\cite{vnect, vibe, smpl, lemo, humor, burenius20133d, cai2019exploiting, dabral2018learning, elhayek2015efficient, elhayek2015outdoor, kiciroglu2020activemocap, liu2020attention, mehta2020xnect, xu2020deep, cliff, fastmetro, nemo, niki, talkshow} can be categorized as end-to-end learning-based or optimization-based approaches. End-to-end methods, such as VNect \cite{vnect}, directly extract 2D and 3D joint positions using CNN-based regression, while VIBE \cite{vibe} estimates SMPL body model~\cite{smpl} parameters using a temporal generation model trained with a discriminator. Other works, such as LEMO \cite{lemo} and HuMoR \cite{humor}, train priors for motion transition using large-scale data \cite{amass, rich, prox, imapper}, which are used for fitting 3D poses from 2d poses extracted by off-the-shelf models~\cite{vitpose,vitpose+} during optimization. However, these methods may produce physically implausible results, such as body twists and foot skating, especially for complex actions or when training data is limited.

\begin{figure*}
    \centering
    \includegraphics[width=\linewidth]{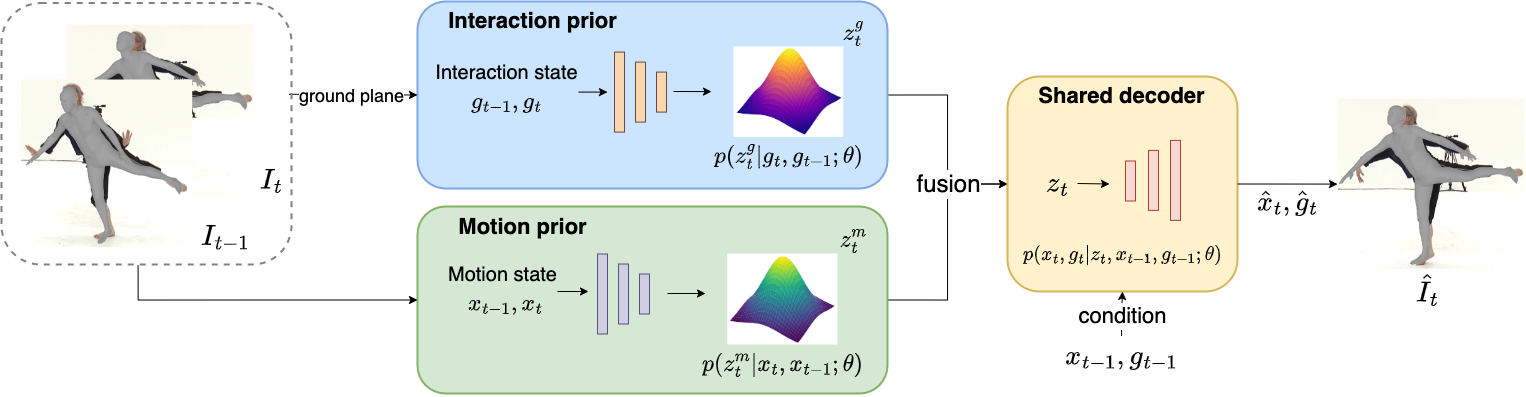}
    \caption{\textit{\bodymethodname architecture}. In training, given the previous state $I_{t-1}$ and current state $I_t$, we obtain the motion state $x_{t-1}$, $x_t$, and interaction state $g_{t-1}$, $g_t$. Our model learns the transition of motion and interaction state changes separately by two priors and reconstructs $\hat{x}_{t}$, $\hat{g}_t$ by sampling from the two distributions and decoding them conditioned on both $x_{t-1}$ and $g_{t-1}$.}
    \label{fig:dual_prior_network}
\end{figure*}

\textbf{Physics-based estimation with simulators.}
Several methods~\cite{dynamics-regulated,simpoe,posetriplet,neuralmocon,diffphy} have been proposed to enhance physical plausibility by incorporating physics laws. These methods use physics simulators such as MuJoCo~\cite{mujoco} and Isaac Gym~\cite{isaacgym} as a black box to guide 3D pose prediction. Due to the non-differentiable nature of the physics simulator, reinforcement learning is employed to learn control of the simulator~\cite{dynamics-regulated,simpoe}. For instance, SimPoE~\cite{simpoe} uses a kinematic-aware policy to generate control signals for the physics simulator to recover realistic 3D poses. Similarly, PoseTriplet~\cite{posetriplet} incorporates the simulator into a semi-supervised framework to reduce artifacts in pseudo labels. Although effective, they can be computationally intensive for training from scratch and prone to collapse, limiting their generalization ability on videos in the wild. To address this issue, differentiable simulators such as TDS~\cite{diffphy} are introduced for articulated 3D motion reconstruction.

\textbf{Physics-based estimation without simulators.} 
Recent research~\cite{physcap, lemo, contact-dynamics, gravicap, kevin-physics} has focused on developing physical constraints for 3D motion optimization that do not require physics simulators~\cite{physcap, lemo, contact-dynamics, gravicap, kevin-physics}. These methods learn to predict contact conditions for specific joints, imposing boundary constraints during optimization.
GraviCap~\cite{gravicap} incorporates the physical properties of moving objects in a scene to recover scale, bone length, and ground simultaneously. However, these constraints are only applied to contact joints and overlook the physical characteristics of the body's other joints. \cite{kevin-physics} infers reaction forces from contact joints and forwards them to the entire body via dynamic equations, but this approach results in approximation errors.
In our work, we propose a continuous representation of human-ground interaction that enables us to investigate interaction conditions for all joints, including non-contact ones.

\textbf{Human-ground interaction representation in pose estimation.}
In physics-based methods for pose estimation~\cite{contact-dynamics,humor,lemo,shimada-neural,ipman}, to impose constraints on height, velocity, and ground reaction forces during optimization, human-ground interaction is typically defined in three ways, the foot-ground contact signal, a contact variable related to penetration distances and contact forces, or the mass center. However, these methods only consider binary contact and ignore non-contact joints. To address this limitation, we propose a continuous and expressive representation for human-ground interaction and establish CVAE-based generative model for human-ground relations to achieve physically plausible motions and reasonable ground planes.

\begin{figure}
    \centering
    \includegraphics[width=\linewidth]{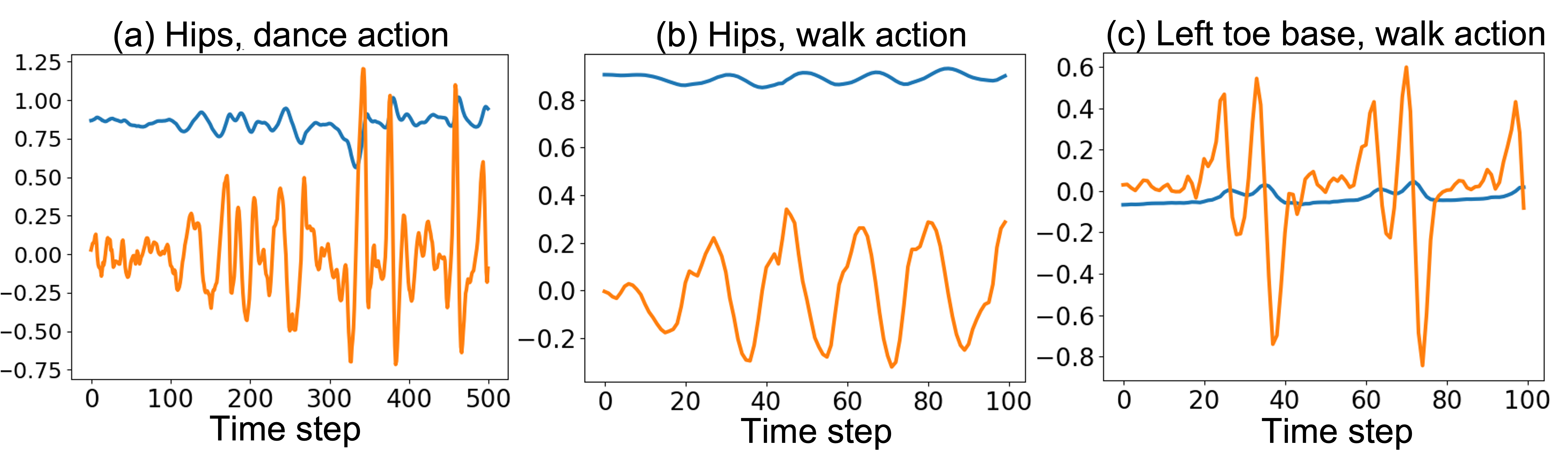}
    \caption{Analysis of the interaction state $g$ defined in Section~\ref{sec: interaction analysis}. We see its components $\textcolor{HaveLockBlue}{d}$ and $\textcolor{orange}{v}$ present unique and dense patterns in separating different types of motion in (a)-(b) and different joints in the same sequence in (b)-(c).}
    \label{fig:motivation}
\end{figure}

\section{Method}
We propose \bodymethodname, a robust generative motion model that captures the dynamics of 3D human motion while being ground-aware, and demonstrate its effectiveness as a regularizer in optimization-based approaches for estimating accurate and plausible 3D human motion and ground plane.

\textbf{Preliminaries.}
With the frame state $I$, we represent the state of a person by an interaction state $g$ defined in the subsequent section, and a motion state $x$ following~\cite{humor}. The motion state $x$ is composed of a root translation $r \in \mathbb{R}^3$, a root orientation $\Phi \in \mathbb{R}^3$, body pose joint angles $\Theta \in \mathbb{R}^{3\times 21}$, joint positions $J \in \mathbb{R}^{3\times 22}$ and their velocities. All the angles are in axis-angle format. 

\subsection{Analysis of Human-ground Interaction} \label{sec: interaction analysis}

\textbf{Representation for human-ground interaction state.}
In contrast to binary contact labels, our objective is to devise a more comprehensive representation that not only indicates whether a joint contacts the ground, but also characterizes the interaction state between joints and the ground in the present, past, and, most importantly, immediate future. This will enable capturing information regarding joints approaching the ground, moving away from the ground, and remaining stationary in the air.

To this end, we represent the human-ground relationship as $g = [d, v]$, consisting of the human-to-ground distance $d\in\mathbb{R}^{23}$ between all joints (including the root joint) and the ground, as well as its velocity $v\in\mathbb{R}^{23}$ along the gravity axis. We employ SMPL body model~\cite{smpl} and utilize the first 23 joints for calculations.

To calculate the human-to-ground distance, we use either the assumed ground or the ground variable $n \in \mathbb{R}^4$ to be optimized at each step, which will be discussed in Section~\ref{sec:joint_optim_strategy}. With the ground plane $n$, we can get a random point $Q$ on it. For the $i$-th joint $J_i$, there is an angle $\alpha_i$ between the ground plane normal $\vec{n}_d \in \mathbb{R}^3$ and vector $\overrightarrow{QJ}_i$. By calculating the projection of vector $\overrightarrow{QJ}_i$ on the plane normal $\vec{n}_d$, we can get the distance representation as follows,
\begin{equation}
\begin{split}
& d^i = |\overrightarrow{QJ}_i| \cdot cos(\alpha_i),\  d = [d^0, d^1, \dots, d^{22}].
\end{split}
\end{equation}

Moreover, we assume that the ground is flat, rigid, and has a floor normal vector oriented along the gravitational axis. In this work, we also make the assumption that the human body primarily interacts with the ground, a circumstance encountered in most in-the-wild cases, such as dance, yoga, and other activities.

\textbf{Analysis of the interaction state.}
As shown in Fig.~\ref{fig:motivation}, our interaction state $g$, including distance $d$ and distance velocity $v$, presents unique and dense patterns in separating different types of motion (Fig.~\ref{fig:motivation}(a)-(b)) and different joints in the same motion sequence (Fig.~\ref{fig:motivation}(b)-(c)).

\textbf{Comparison with binary contact label.} 
Compared to the binary contact label, the continuous interaction representation as shown in Fig.~\ref{fig:motivation} provides more detailed information beyond mere contact. For example, suppose the distance between a joint and the ground is zero, and the velocity along the gravitational direction is significant. It indicates that the joint has recently made contact with the ground, and due to inertia, both the joint and its adjacent counterparts are likely to continue moving toward the ground for the next few frames. Under these conditions, it is highly improbable for the joints to exhibit any motion in the opposite direction.

\begin{figure}
    \centering
    \includegraphics[width=\linewidth]{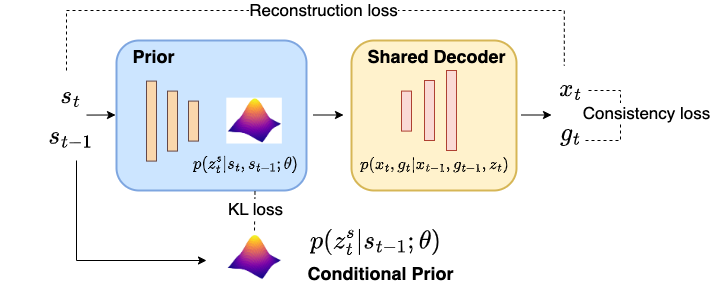}
    \caption{Training of \bodymethodname. For simplicity, ``Prior'' can be either interaction prior or motion prior. Similarly, $s_t$ can indicate $g_t$ and $x_t$, depending on the prior type.}
    \label{fig:dual_prior_training}
\end{figure}

\subsection{Ground-aware Generative Motion Model}
Building upon the proposed representation, we aim to develop an explicit ground-aware motion dynamics model that incorporates human-ground interactions with human pose to capture the temporal variations in human pose and human-ground interactions. 

Specifically, we model the probability of a motion sequence $x_t$ by considering the human-ground interaction $g_t$ at each step, \textit{i.e.},
\begin{equation}
p_\theta (x_0, g_0, x_1, g_1, \cdots, x_T, g_T) = p_\theta(x_0,g_0) \prod_{t=1}^{T}p_\theta(x_t, g_t|x_{t-1}, g_{t-1}),
\end{equation}
where $x_t$ and $g_t$ are the motion and interaction states at the time step, respectively.
For each time step, the overall motion depends not only on the motion state $x_{t-1}$ at the previous time step but also on the human-ground interaction state formulated as $g_{t-1}$. Consequently, this allows $p(x_t, g_t|x_{t-1}, g_{t-1})$ to capture the fine-grained physical plausibility of the transition.

As illustrated in Fig.~\ref{fig:dual_prior_network}, we propose \bodymethodname that leverages a conditional variational autoencoder (CVAE) to model the transition probability. This model formulates the probability of transition in motion state and interaction state as follows:
\begin{equation}
p_\theta (x_t, g_t|x_{t-1}, g_{t-1}) = p_\theta (z_t|x_{t-1}, g_{t-1}) \cdot p_\theta (x_t, g_t|z_t),
\end{equation}
where $z_t$ is the latent variable for time step $t$. For the purpose of computation efficiency, we formulate $p_\theta (z_t|x_{t-1}, g_{t-1})$ into two independent probabilities as:
\begin{equation}
p_\theta (z_t^{m} | x_{t-1}), p_\theta (z_t^{g} | g_{t-1}),\ \mathrm{ s.t. }\ z_t = z_t^{m} \oplus z_t^{g},
\end{equation}
where $z_t^m$ and $z_t^g$ denote the latent transitions for motion and human-ground interaction respectively, and $\oplus$ denotes the concatenation operation in implementation. During training, these two probabilities are learned by two priors with the adjacent states as input, instead of the previous state. They are approximated as independent Gaussian distributions using implicit neural networks. 
\begin{equation}
p_\theta (z^m_t | x_{t-1}, x_t), \ p_\theta (z^g_t | g_{t-1}, g_t)
\end{equation}
To enable the differentiation and learning of unique characteristics for the two priors, we employ two conditional priors as guidance rather than relying on the standard Gaussian distribution, \textit{i.e.},
\begin{equation}
\begin{split}
& p_\theta (z^{m}_t | x_{t-1}) = \mathcal{N} (z^{m}_t; \mu_\theta(x_{t-1}), \sigma_\theta(x_{t-1})), \\
& p_\theta (z^{g}_t | g_{t-1}) = \mathcal{N} (z^{g}_t; \mu_\theta(g_{t-1}), \sigma_\theta(g_{t-1})).
\end{split}
\end{equation}
By simultaneously learning the distribution of transitions for both pose and joint-to-ground interactions across adjacent frames within a motion sequence, our model can produce a wide range of plausible poses while being ground-aware. 

In the next step, we employ \textbf{a shared decoder} to estimate the future motion conditioned on both the motion state and the human-ground interaction from the previous step, thereby ensuring consistency between the body pose and the ground plane. Specifically, the shared decoder is designed to enable the combination of multiple inputs, including a random motion latent sample $z_t^m$, a random interaction latent sample $z_t^g$ (with the combined latent variables denoted as $z_t$), the motion state $x_{t-1}$, and the interaction representation $g_{t-1}$. Besides, to facilitate an auto-regressive manner in further applications, it outputs the motion state $x_t$ and the interaction $g_t$ simultaneously. Similar to the baseline, it also predicts a binary contact label $c_t$ for the predefined nine contact joints. 

\textbf{Training loss and implementation details.}
As in Fig.~\ref{fig:dual_prior_training}, the training loss contains reconstruction loss $\mathcal{L}_{recon}$ for motion state and interaction state, KL loss $\mathcal{L}_{KL}$ between conditional prior and the corresponding encoder output, and consistency loss $\mathcal{L}_{consist}$ between motion state and learned interaction state, \textit{i.e.},
\begin{equation}
\mathcal{L} = \mathcal{L}_{recon} + \mathcal{L}_{KL} + \mathcal{L}_{consist},
\end{equation}
where the reconstruction loss $\mathcal{L}_{recon}$ is defined as:
\begin{equation}
\mathcal{L}_{recon} = ||x_t - \hat{x}_t ||^2 + ||g_t - \hat{g}_t ||^2,
\end{equation}
given the training pair $(x_t, g_t, x_{t-1}, g_{t-1})$. $\hat{x}_t$ and $\hat{g}_t$ are the output of the decoder for motion state and interaction state, respectively. The KL loss $\mathcal{L}_{KL}$ is calculated separately for motion and interaction states by computing the KL divergence $D_{KL}(\cdot || \cdot)$ between the output of the encoder and the corresponding conditional prior. The consistency loss $\mathcal{L}_{consist}$ promotes consistency between the learned interaction state $\hat{g}_t$ and the human-ground interaction information, which is extracted through the function $f(\cdot)$ from the predicted joints $\hat{x}_t$ and the ground truth ground plane $n$, \textit{i.e.},
\begin{equation}
\mathcal{L}_{consist} = ||\hat{g}_t - f(\hat{x}_t, n) ||^2.
\end{equation}

Lastly, for comparing the contact accuracy with the baseline, we also incorporate a contact classification head and compute the BCE loss between the predicted contact label $\hat{c}_t$ and the ground truth.

\subsection{Joint Optimization Strategy}\label{sec:joint_optim_strategy}
After training, following \cite{humor}, we devise a joint optimization strategy for 3D human motion reconstruction from noisy observations and RGB videos. We leverage \bodymethodname to regularize optimization toward the space of plausible ground-aware motions, thereby maintaining consistency between the human body and the ground plane. We consider our \bodymethodname for two tasks: (1) denoising under the fixed ground plane; and (2) motion reconstruction from RGB videos where the ground plane is unavailable and subject to optimization alongside the motion sequence.

\textbf{Optimization variables.}
Given a sequence of motion observations $y_{0:T}$ in 2D/3D joints format and an optional ground plane $n$, we aim to obtain a sequence of SMPL parameters $(r_{0:T}, \Phi_{0:T}, \Theta_{0:T})$, body shape $\beta$, and ground plane $n$ (if not provided), which could not only match the observation but also maintain physical plausibility and consistency between human and ground. Our \bodymethodname could be incorporated into the optimization by parameterizing the SMPL parameter sequence into an initial motion state $x_0$, an initial interaction state $g_0$, and a sequence of latent variables $z_{1:T}$ composed of motion latent variables $z_{1:T}^m$ and interaction latent variables $z_{1:T}^g$. With optimized latent variables and initial states, we can roll-out the whole sequence of SMPL parameters through the decoder in an auto-regressive way.

\textbf{Noisy observation setting.}
In this setting, we consider a scenario where a ground plane and a set of joint positions, generated using existing motion reconstruction algorithms like SMPLify~\cite{smplify}, are available in a noisy form. Our goal is to optimize the motion sequence to ensure both physical plausibility and accuracy in human-ground interactions when the ground plane is provided and fixed. We show our model performs better when used for fitting to noisy joints and known ground planes, especially in challenging cases.

To this end, the objective function is formulated as a combination of dual-prior loss, prior consistency loss, data loss, and regularization loss, with the last two loss terms following the design of \cite{humor}. In this context, we primarily focus on the dual-prior loss:
\begin{equation}
\begin{split}
L_{prior} & = \prod_{t=1}^T log \mathcal{N} (z_t^{m}; \mu_\theta (x_{t-1}), \sigma_\theta (x_{t-1})) \\
& + \prod_{t=1}^T log \mathcal{N} (z_t^{g}; \mu_\theta (g_{t-1}), \sigma_\theta (g_{t-1})),
\end{split}
\end{equation}
and the prior consistency loss:
\begin{equation}
L_{pconsist} = \prod_{t=1}^T || g_t - f(x_t, n) ||^2,
\label{eq:loss_pconsist}
\end{equation}
where $L_{prior}$ adopts the learned conditional priors for calculation. $n$ is the fixed ground plane normal.

\textbf{RGB video setting.}
In this particular setting, we tackle the problem of motion reconstruction from RGB videos, where a set of 2D/3D keypoint positions $y_{0:T}$, extracted from each individual frame in the camera view, is provided, but without any knowledge of the ground plane. Our objective is to seek both the physically plausible and precise motion state $x_{0:T}$ and the ground plane $n$ that can transform the motion state into world space.

In contrast to the noisy observation setting, the ground plane is unavailable here. As a result, we establish a ground plane variable $n$, allowing it to be optimized alongside the motion state. In total, we optimize the motion and interaction latent variables $z_{1:T}^m$, $z_{1:T}^g$, initial motion and interaction states $x_0$, $g_0$, and the ground plane vector $n$ at the same time. In each optimization iteration, the prior consistency loss, as shown in Eq.~\eqref{eq:loss_pconsist}, is calculated based on the optimized ground plane vector $n$ rather than the assumed one.

\textbf{Implementation details}.
During optimization, initialization phase includes latent variables $(z_{1:T}^m, z_{1:T}^g)$ and first-frame motion and interaction states $x_0$, $g_0$. We first initialize the SMPL parameters by a single-frame algorithm~\cite{smplify,pare}, and thus obtain the initialization of first frame states $x_0$, $g_0$ and the latent variables $(z_{1:T}^m, z_{1:T}^g)$ through the trained priors $p_\theta (z_t^m | x_{t-1}, x_t)$, $p_\theta (z_t^g | g_{t-1}, g_t)$. 

\section{Experiment}

\subsection{Datasets and Splits}
\textbf{AMASS}~\cite{amass} is a large motion capture dataset containing multiple types of motions, mainly running, walking, and turning around. We follow \cite{humor} to process the sequences into 30 hz and extract the contact labels for evaluation. Our model and the baseline are both trained on the training set of AMASS and evaluated on the test set of all datasets without retraining or fine-tuning. 

To assess the effectiveness of our proposed model in handling various types of human-ground relationships, we partition the test set of AMASS into distinct levels according to the minimum hip height within each sequence. Our hypothesis is that as the minimum hip height decreases, the interaction between the human and the ground becomes increasingly intricate.

\textbf{AIST++} dataset~\cite{aistpp} comprises a vast collection of dance motion data that includes RGB videos, multiple camera parameters, and 3D motion annotations for 1,408 sequences of 30 different subjects. For the purpose of evaluating our model's performance under different human-ground relations, we partition the test set according to the degree of difficulty involved in estimating the ground plane. 

\begin{table*}
\begin{center}
\begin{tabular}{l|cccccc}
\hline
Method & MPJPE-G ($\downarrow$) & MPJPE ($\downarrow$) & MPJPE-PA ($\downarrow$) & PVE ($\downarrow$) & contact acc ($\uparrow$) & accel mag ($\downarrow$) \\
\hline
VPoser-t~\cite{vposer} & 32.8 & 34.8 & 27.9 & 43.2 & - & 61.6 \\
VPoser-t + HuMoR~\cite{humor} &  22.7 & 23.9 & 19.0 & 30.3 & 89.3\% & \textbf{16.7} \\
VPoser-t + \textbf{\bodymethodname} & \textbf{21.9} & \textbf{22.6} & \textbf{18.1} & \textbf{29.5} & \textbf{91.1\%} & 20.8 \\
\hline
\end{tabular}
\end{center}
\caption{Results on the AMASS dataset under the noisy observation setting.}
\label{tab:known_ground_setting_results}
\end{table*}

\begin{figure*}
    \centering
    \includegraphics[width=\linewidth]{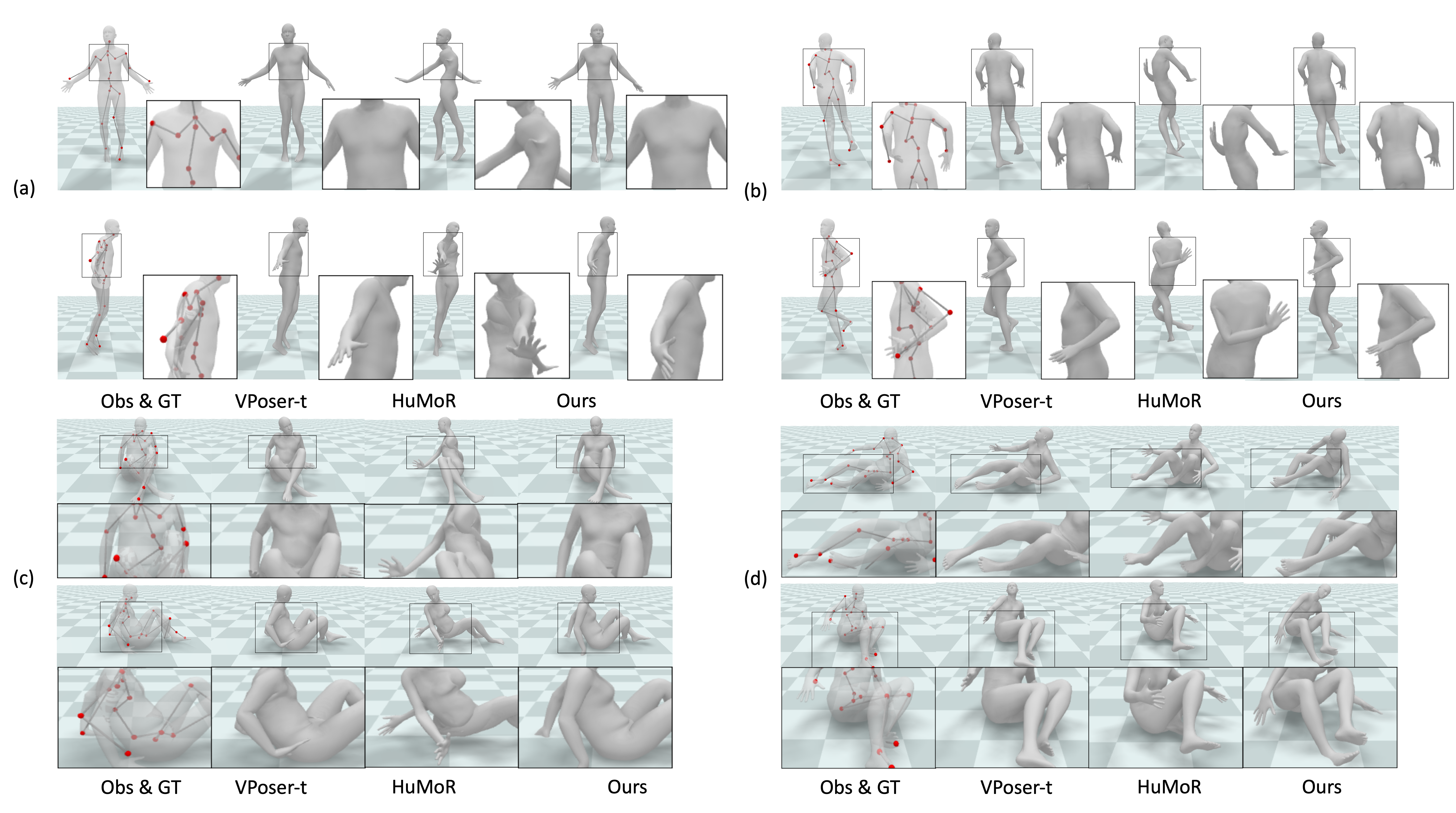}
    \caption{Qualitative comparison on the AMASS test set under the noisy observation setting. Our method doesn't show body twist even under complex human-ground interaction. For each case, the first row shows the front view in the world space, while the second row shows the side view. Please view the supplementary video for more details.}
    \label{fig:qual_comp_known_ground}
\end{figure*}

\subsection{Baselines and Metrics}
\textbf{Baselines.} We conducted a comparative analysis between our method and the baseline HuMoR~\cite{humor}, a CVAE-based prior that does not take dense human-ground interaction into account. We ensure that the initialization and optimization settings are identical for both methods. In the noisy observation setting, VPoser-t serves as the initialization algorithm, while in the RGB video setting, we use PARE~\cite{pare}, a single-frame learning-based pose reconstruction technique, for initialization. VPoser-t uses VPoser~\cite{vposer} and 3D joints smoothness constraints during optimization.

\textbf{Metrics.} In our evaluation, we employ several common metrics to assess the performance of our method and the baseline. The 3D positional errors are measured by the mean per joint position error (MPJPE), MPJPE after procrustes alignment (MPJPE-PA), MPJPE over global positions (MPJPE-G), and the per-vertex error (PVE). In addition, we evaluate the binary classification accuracy of nine pre-defined joints~\cite{humor} that are likely to be in direct contact with the ground. We also assess the smoothness of the generated motion by computing the average per-joint accelerations (Accel). Moreover, we report the performance of our method on different levels of human-ground interaction, which cannot be captured by the overall errors on the entire test set. We also report the cosine similarity scores (Cos) between normal vectors of planes to evaluate performance in estimating the ground plane.

\subsection{Optimization with noisy observations}
First, we evaluate \bodymethodname with the observation of noisy 3D joint positions and a fixed ground plane, and demonstrate that \bodymethodname performs better than the baseline, especially in cases with complex human-ground relations. We use the 90-frame (3s) clips from the AMASS dataset. To simulate the presence of noise, we introduce Gaussian noise to the joint positions with a mean of zero and a standard deviation of 0.04m, following~\cite{humor}.

Table~\ref{tab:known_ground_setting_results} presents the mean results attained over the entire test set of the AMASS dataset. We compare \bodymethodname with baseline HuMoR, as well as the initialization method VPoser-t. Our results demonstrate that our \bodymethodname approach produces more precise poses and yields better performance in terms of contact accuracy. These findings suggest that the use of interaction states facilitates the extraction of human-ground interaction and significantly enhances human-ground relations. Regarding smoothness, while HuMoR reports the lowest acceleration, our approach outperforms VPoser-t substantially and provides an inherently smooth outcome. In contrast to HuMoR, our method affords greater flexibility to accommodate noisy poses, particularly those characterized by complex human-ground relations.

\begin{figure}
    \centering
    \includegraphics[width=0.92\linewidth]{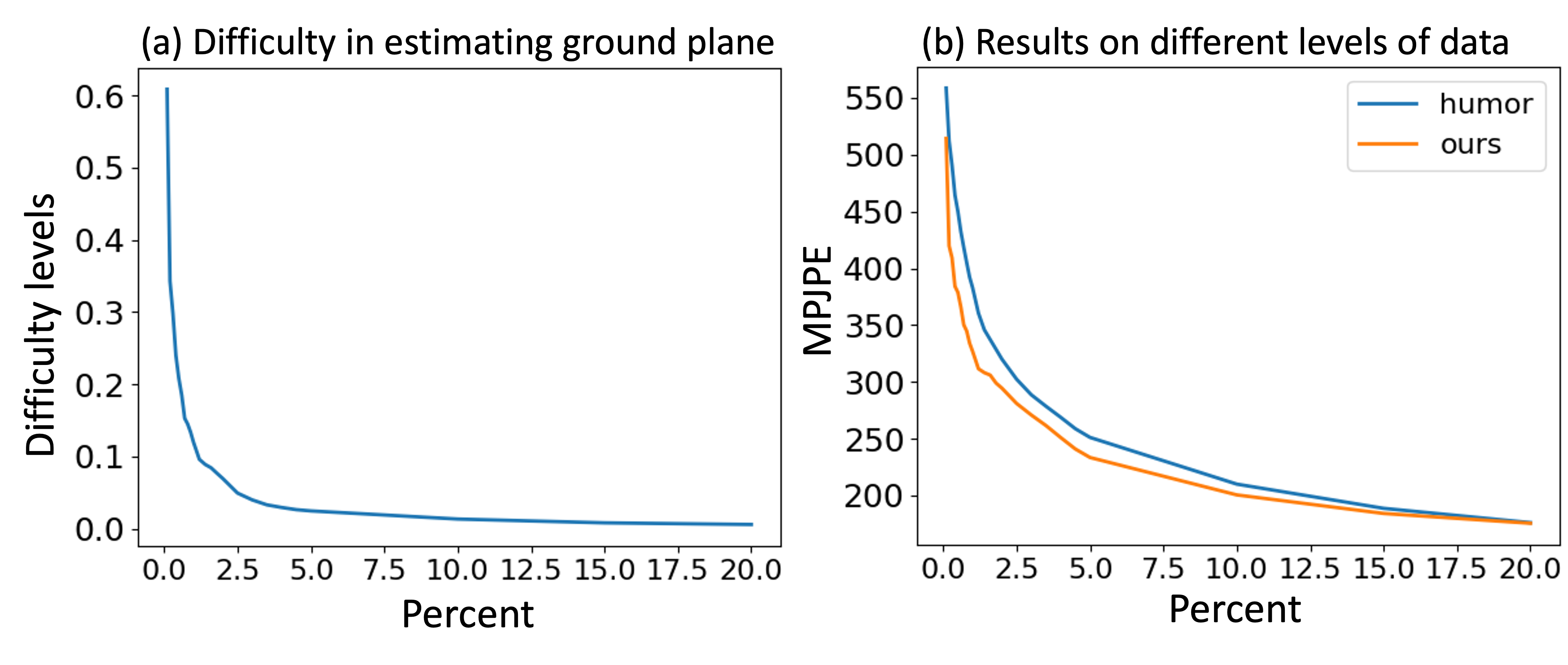}
    \caption{(a) Difficulty levels of the top 20\% of challenging data. (b) Our method outperforms the baseline HuMoR in the top 20\% of challenging cases.}
    \label{fig:aistpp_challenging_cases}
\end{figure}

\begin{table}
\begin{center}
\begin{tabular}{l|l|cccc}
\hline
Method & Metric & 0-0.3 & 0.3-0.6 & 0.6-1.0 & avg  \\
\hline
\multirow{3}{4em}{VPoser-t \cite{vposer}} & MPJPE-G ($\downarrow$) & \textbf{34.4} & 33.5 & 32.6 & 33.5\\
& MPJPE ($\downarrow$) & \textbf{38.0} & 36.0 & 34.5 & 36.2 \\
& PVE ($\downarrow$) & \textbf{48.1} & 44.4 & 43.0 & 45.2  \\
\hline
\multirow{4}{4em}{VPoser-t + HuMoR \cite{humor}} & MPJPE-G ($\downarrow$) & 53.0 & 24.0 & 21.6 & 32.9\\
& MPJPE ($\downarrow$) & 53.0 & 25.8 & 22.8 & 33.9 \\
& PVE ($\downarrow$) & 65.3 & 33.1 & 28.9 & 42.4  \\
& contact acc ($\uparrow$) & 75.9\% & 86.6\% & 90.0\% & 84.2\% \\
\hline
\multirow{4}{4em}{VPoser-t + \textbf{\bodymethodname}} & MPJPE-G ($\downarrow$) & 48.2 & \textbf{22.6} & \textbf{21.1} & \textbf{30.6} \\
& MPJPE ($\downarrow$) & 49.1 & \textbf{23.9} & \textbf{21.7} & \textbf{31.6} \\
& PVE ($\downarrow$) & 59.6 & \textbf{30.4} & \textbf{28.5} & \textbf{39.5} \\
& contact acc ($\uparrow$) & \textbf{78.6\%} & \textbf{87.6\%} & \textbf{91.8\%} & \textbf{86.0\%} \\
\hline
\end{tabular}
\end{center}
\caption{Results on AMASS dataset under the noisy observation setting at different levels of human-ground interaction.}
\label{tab:amass_eval_by_distance}
\end{table}

\begin{table*}
\begin{center}
\begin{tabular}{l|ccccccccc}
\hline
Method & Cos ($\uparrow$) & Cos 1\% ($\uparrow$) & Accel ($\downarrow$) & Accel align ($\downarrow$) & MPJPE-G ($\downarrow$) & MPJPE ($\downarrow$) & MPJPE-PA ($\downarrow$) & MPJPE* 1\% ($\downarrow$)\\
\hline
PARE~\cite{pare} & - & - & 65.6 & 23.8 & \textbf{257.3} & \textbf{102.5} & \textbf{62.0} & - \\
PARE + HuMoR~\cite{humor} & 0.99175 & 0.70452 & \textbf{4.0} & \textbf{3.3} & 606.2 & 114.3 & 80.7 & 383.0 \\
PARE + \textbf{\bodymethodname} & \textbf{0.99965} & \textbf{0.99956} & 4.4 & 3.6 & 666.0 & 130.5 & 92.9 & \textbf{327.0}\\
\hline
\end{tabular}
\end{center}
\caption{Results on AIST++ dataset under the RGB video setting. ``Cos'' is the mean cosine similarity between the predicted ground plane and the gt. ``Cos 1\%'' is the Cos scores of the top 1\% difficult clips in estimating the ground plane. ``MPJPE* 1\%'' denotes the MPJPE of the predictions in world space for the top 1\% difficult data in estimating the ground plane.}
\label{tab:unknown_ground_setting_results}
\end{table*}

Table~\ref{tab:amass_eval_by_distance} presents the outcomes for data splits categorized by varying levels of human-ground interaction. Compared with HuMoR, our approach demonstrates superior performance, particularly in the most challenging level ``0-0.3''. At this level, our method exhibits improvements in both positional error and contact accuracy, indicating that it produces a more physically realistic and accurate pose with a more reasonable contact condition. While VPoser-t displays a consistently robust performance across all levels of data, it is unable to predict the ground plane and exhibits inferior smoothness capabilities. Notably, our method outperforms VPoser-t at ``0.3-0.6'' and ``0.6-1.0'' levels.

Fig.~\ref{fig:qual_comp_known_ground} presents qualitative examples of our approach compared to HuMoR. In Figures~\ref{fig:qual_comp_known_ground}(a) and \ref{fig:qual_comp_known_ground}(b), HuMoR exhibits body twists in jumping, while our method doesn't. As HuMoR lacks an understanding of human-ground interaction, it struggles to accurately discern the motion and distinguish between joint position changes caused by noise versus those caused by the actual action itself. In Figures \ref{fig:qual_comp_known_ground}(c) and \ref{fig:qual_comp_known_ground}(d), HuMoR shows inaccurate orientation and body twist in sitting because of complexity in motion and the human-ground interaction, while our method performs well in these cases.

\subsection{Optimization with RGB video}
Next, we show that our \bodymethodname can predict a more physically reasonable pose and ground plane simultaneously, and can accurately figure out the ambiguous pose under camera view. In this setting, we use 60-frame (2s) video clips from the AIST++ dataset. 

To quantify the challenge of estimating the ground plane for video clips, we assume that a larger divergence in the predicted ground planes by different methods indicates a higher level of difficulty in pose ambiguity. This is due to the fact that the pose in camera view may not readily differentiate the ground plane. To assess this, we calculate the cosine similarity scores of the predicted ground plane from our approach and the baseline HuMoR separately, and then sort clips according to the absolute difference in similarity scores between the two methods. The absolute difference for the top 20\% of clips is shown in Fig.~\ref{fig:aistpp_challenging_cases}(a), while the remaining clips indicate a negligible difference and are therefore not presented.

\begin{figure*}
    \centering
    \includegraphics[width=\linewidth]{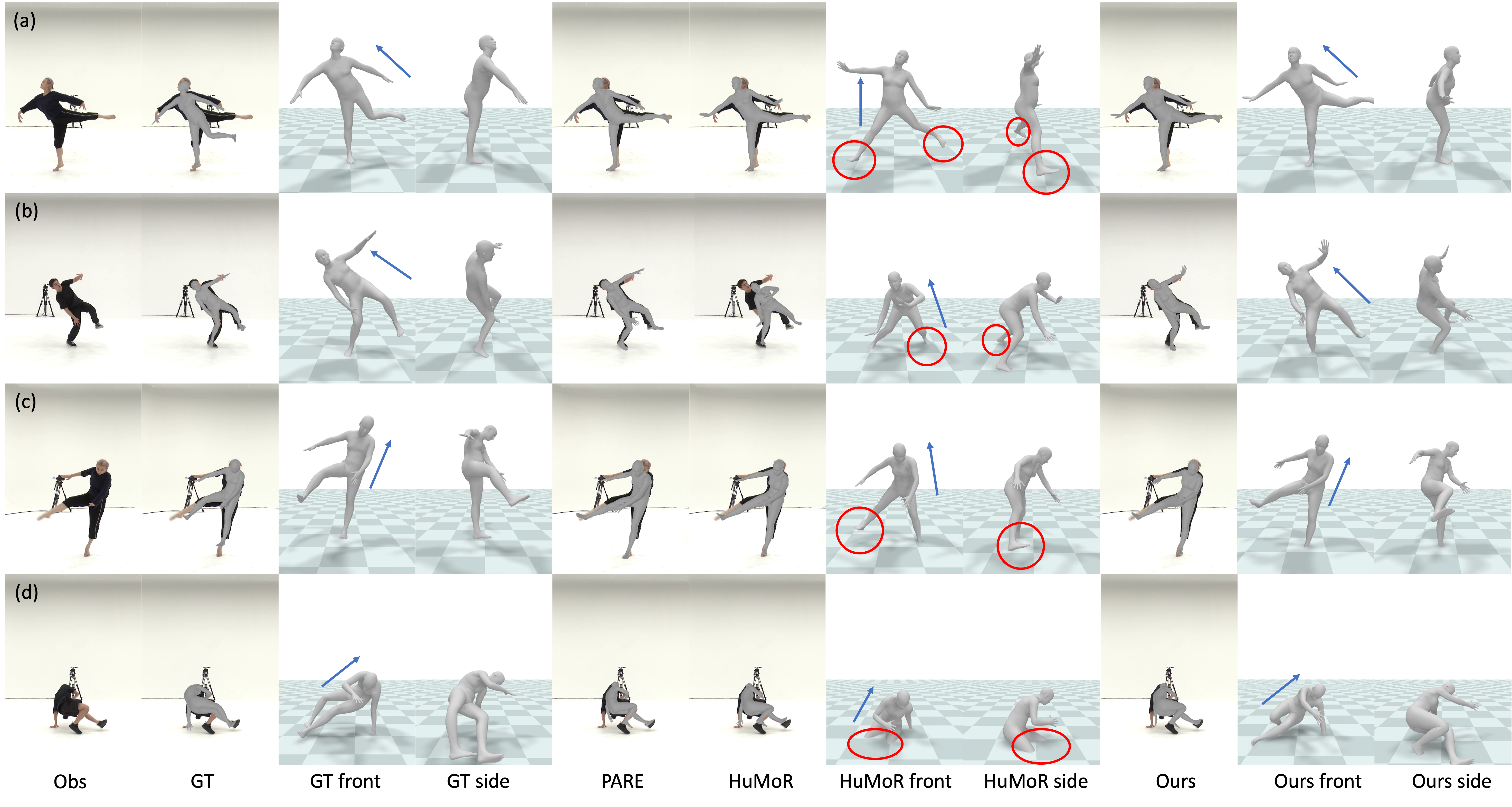}
    \caption{Qualitative comparison on the AIST++ test set under the RGB video setting. ``front'' and ``side'' denote the front and side view in world space. The \textcolor{HaveLockBlue}{direction of the body torso} and \textcolor{red}{contacts of HuMoR} are highlighted. HuMoR tends to predict the body torso in a direction perpendicular to the ground while our method doesn't.} 
    \label{fig:qual_comp_unknown_ground}
\end{figure*}

Table~\ref{tab:unknown_ground_setting_results} presents the results of our method, baseline HuMoR, and the initialization method PARE, for the entire test set and the top 1\% of clips. \bodymethodname exhibits superior performance in estimating the ground plane, particularly for the top 1\% of difficult data regarding human-ground relations. This suggests \bodymethodname can better distinguish between mistaken poses under camera view. In terms of smoothness, both HuMoR and \bodymethodname show significant improvements compared to PARE, albeit at an acceptable cost of position accuracy. Although \bodymethodname reports relatively inferior results regarding positional metrics for the entire test set, it can produce more reasonable poses under the world space by considering the ground plane, especially for ambiguous motions. As shown in Fig.~\ref{fig:aistpp_challenging_cases}(b), regarding the MPJPE in world space, our \bodymethodname outperforms HuMoR on the top 20\% difficult cases. 

The qualitative examples also provide evidence to support this conclusion. Fig.~\ref{fig:qual_comp_unknown_ground} presents some examples from AIST++. We showcase both the prediction in camera view and in world view.
Since PARE cannot predict the ground plane, we exclude its prediction under the world view. As demonstrated in Fig.~\ref{fig:qual_comp_unknown_ground}, HuMoR generates accurate poses in most cases under camera view, but produces completely physically unreasonable poses with incorrect contact conditions in world space.
This suggests that HuMoR is incapable of resolving ambiguous poses in world space and solely optimizes motion by observation. On the other hand, our method, aided by the interaction map, accurately resolves ambiguous poses and generates physically plausible poses with the correct conditions.

\begin{figure}
    \centering
    \includegraphics[width=\linewidth]{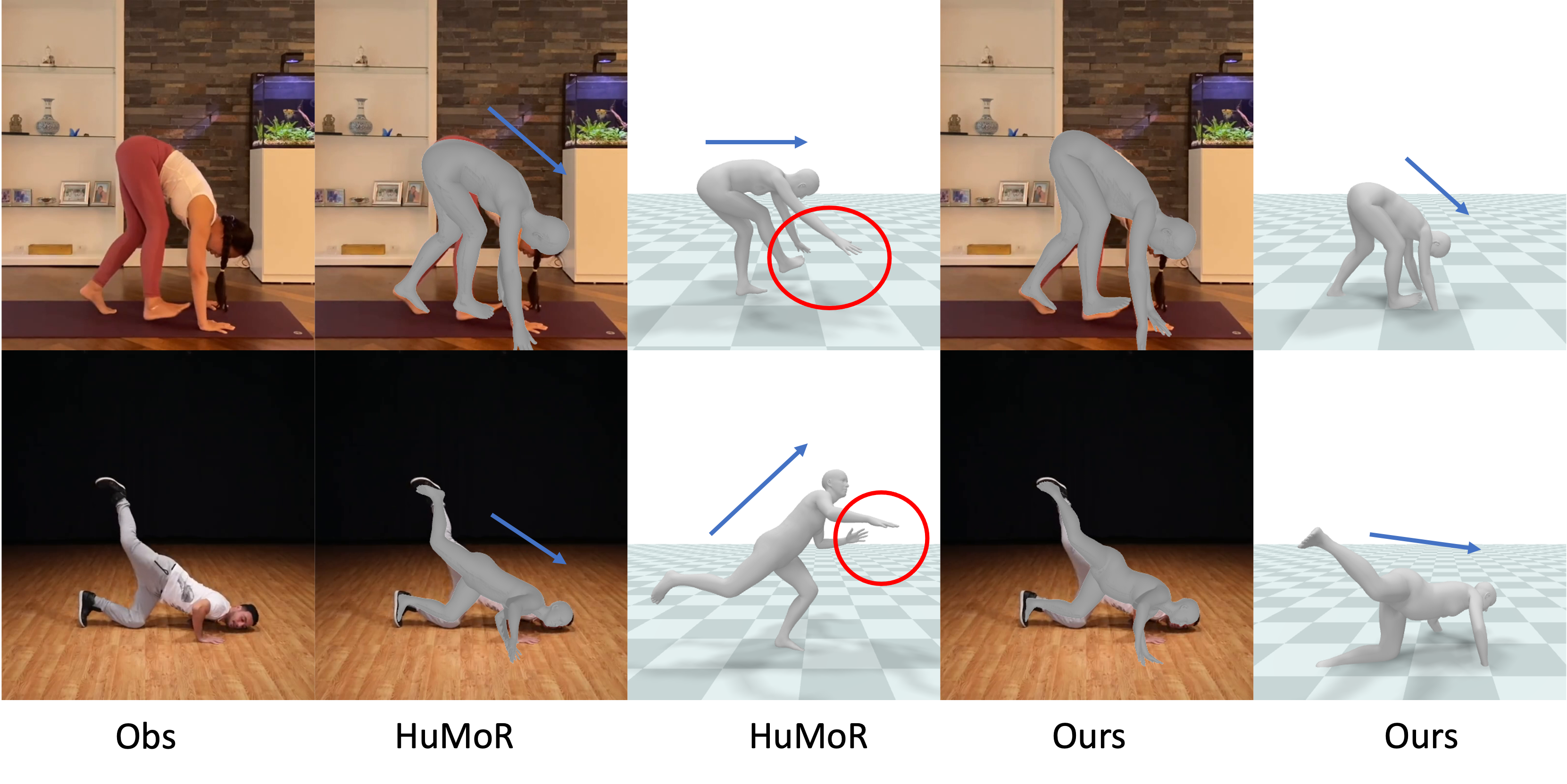}
    \caption{Qualitative comparison on the videos from the Internet. The \textcolor{HaveLockBlue}{blue arrow} and \textcolor{red}{red circle} have the same meaning as in the above figures.}
    \label{fig:internet_examples}
\end{figure}

\textbf{Generalizing to videos in the wild.}
Finally, we compare our method with the baseline HuMoR on videos sourced from the Internet and demonstrate that our method generalizes better to videos in the wild without the need for retraining. Fig.~\ref{fig:internet_examples} showcases the challenging scenarios like yoga, and handstanding.

\section{Limitation and Future Work}
Although our model can yield superior performance in predicting physically plausible motion and reasonable ground planes in challenging cases, there are some limitations, such as inconsistency in hand motion. In some extreme cases, our method can make a reasonable inference on the ground plane but have a large error in positions due to the extreme angle and the high moving speed. Nonetheless, our approach outperforms the baseline method HuMoR in these challenging cases. In future work, it is promising to learn a stronger prior from large-scale training data (\textit{e.g.}, flexible contact joints, fine-grained hand motion) to further improve the performance. More discussion is in Section~\ref{supp_sec:limitation} in the Appendix.

\section{Conclusion}
In this work, we propose a dense and continuous representation for human-ground interaction and a CVAE-based model named \bodymethodname based on it to address the consistency issue between the human and the ground. We further establish a joint optimization-based approach that uses our proposed \bodymethodname as a regularizer to estimate physically plausible and correct motion from noisy observations and RGB videos. The proposed method demonstrates promising results in generating realistic outcomes, particularly in challenging scenarios characterized by complex and ambiguous human-ground interaction.

\newpage
\bibliographystyle{ACM-Reference-Format}
\balance
\bibliography{main}

\newpage
\appendix
\section{Method Details}\label{supp_sec:method}

\subsection{Preliminary}
In this section, we use the same symbols as in \cite{cvae}. $q_\phi (z_t^m|x_{t-1}, x_t)$, $q_\phi (z_t^g|g_{t-1}, g_t)$ indicate the motion and interaction encoders (called motion/interaction prior in Fig.~\ref{fig:dual_prior_network} of main paper). $p_\theta (z_t^m|x_{t-1})$, $p_\theta (z_t^g|g_{t-1})$ are the conditional priors for motion and interaction respectively, as shown in Fig.~\ref{fig:dual_prior_training} in the main paper. 

\subsection{GraMMaR: Ground-aware Motion Model}

\noindent\textbf{Formulation.}
From a probabilistic perspective, our final goal is to model the probability of a sequence of physically plausible motion states with reasonable human-ground relations,
\begin{equation}
p_\theta(x_0,g_0,x_1,g_1,\cdots,x_T,g_T) = p_\theta(x_0,g_0) \prod_{t=1}^T p_\theta(x_t,g_t|x_{t-1},g_{t-1})
\end{equation}
where each motion state $x_t$ combined with the interaction state $g_t$ is only dependent on the previous states $x_{t-1}, g_{t-1}$.

To achieve it, we propose a CVAE-based model that learns the probability of the transition of motion states and the interaction states as follows,
\begin{equation}
p_\theta(x_t, g_t | x_{t-1}, g_{t-1}) = \int_{z_t} p_\theta (z_t|x_{t-1},g_{t-1}) p_\theta (x_t,g_t|z_t,x_{t-1},g_{t-1})
\label{eq:transition_prob_v1}
\end{equation}
where $z_t$ is the latent variables. For the purpose of computation efficiency, we assume that latent variables $z_t$ consist of two independent components $z_t^m$ as motion latent variables and $z_t^g$ as interaction latent variables. 
\begin{equation}
z_t = z_t^m \oplus z_t^g,
\end{equation}
where $\oplus$ is the concatenation operation in implementation. Therefore, we can reformulated Eq.~\ref{eq:transition_prob_v1} as
\begin{equation}
\begin{split}
& p_\theta(x_t, g_t | x_{t-1}, g_{t-1})  \\
& = \int_{z_t} p_\theta (z_t^m | x_{t-1}) p_\theta (z_t^g | g_{t-1}) p_\theta (x_t,g_t|z_t,x_{t-1},g_{t-1}),\\
& s.t. z_t = z_t^m \oplus z_t^g.
\end{split}
\end{equation}

\noindent\textbf{Architecture.}
\begin{figure*}
    \centering
    \includegraphics[width=\linewidth]{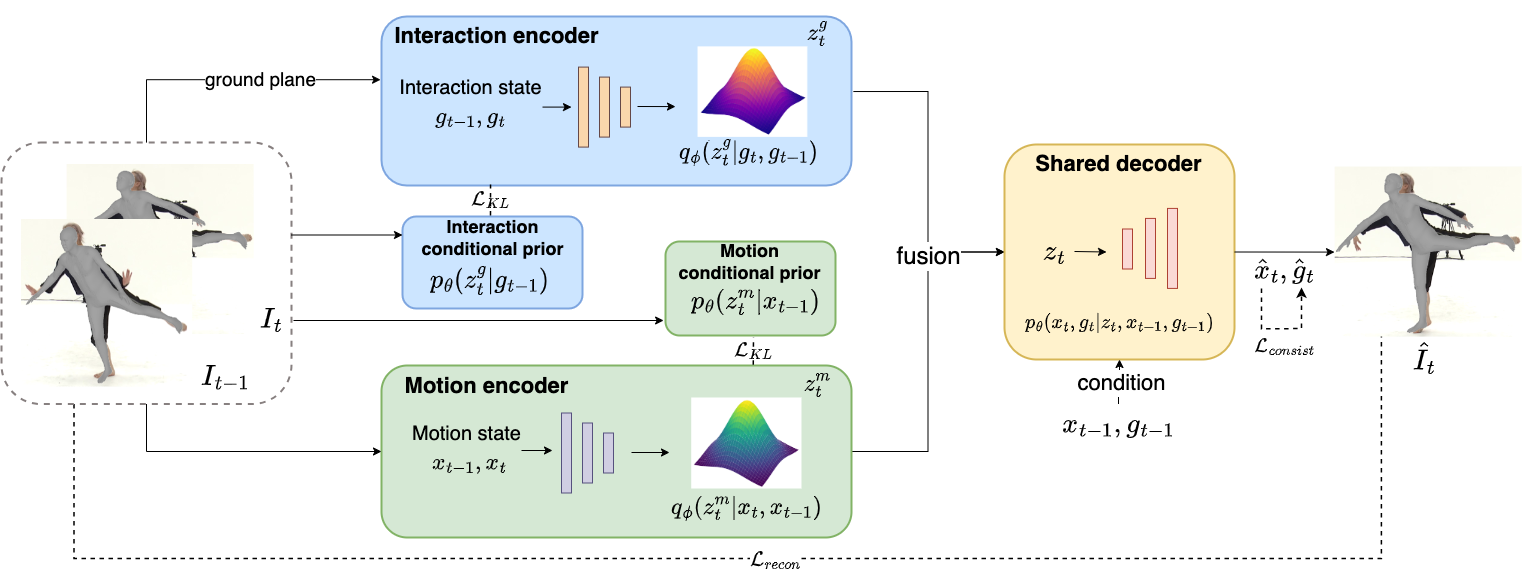}
    \caption{\bodymethodname architecture. In training, given the previous state $I_{t-1}$ and current state $I_t$, we obtain the motion state $x_{t-1}$, $x_t$, and interaction state $g_{t-1}$, $g_t$. Our model learns the transition of motion and interaction state changes separately by two priors and reconstructs $\hat{x}_{t}$, $\hat{g}_t$ by sampling from the two distributions and decoding them conditioned on both $x_{t-1}$ and $g_{t-1}$.}
    \label{supp_fig:network_complete}
\end{figure*}
As shown in Fig.~\ref{supp_fig:network_complete}, we present \bodymethodname, an explicit ground-aware motion dynamics model that incorporates human-ground interactions and human poses. It is formulated as a conditional variational auto-encoder based model. Specifically, we propose two separate encoders to simultaneously learn the probabilities of transition for both pose and joint-to-ground distances across adjacent frames within a motion sequence, one for interaction transition $q_\phi (z_t^g | g_{t-1}, g_t)$, another for motion transition $q_\phi (z_t^m | x_{t-1}, x_t)$, producing a wide range of plausible poses and human-ground interactions.
Similarly, there are two conditional priors to model the motion and interaction distribution $p_\theta (z_t^m | x_{t-1})$ and $p_\theta (z_t^g | g_{t-1})$.

Finally, to model $p_\theta (x_t, g_t | z_t, x_{t-1}, g_{t-1})$, we adopt a shared decoder that generates the motion and interaction states $x_t$, $g_t$ conditioned on samples $z_t$ from the two learned distributions and the previous states $x_{t-1}$, $g_{t-1}$. We can either use the conditional priors $p_\theta (z_t^m | x_{t-1})$, $p_\theta (z_t^g | g_{t-1})$ or the encoders $q_\phi (z_t^m | x_{t-1}, x_t)$, $p_\phi (z_t^g | g_{t-1}, g_t)$ to generate the latent variables under different scenarios, which depends on the input types.

\noindent\textbf{Training details.}
Our \bodymethodname model is trained to approach the lower bound as follows,
\begin{equation}
\begin{split}
& \log p_\theta (x_t, g_t | x_{t-1}, g_{t-1}) \\
& \geq \mathbb{E}_{q_\phi (z_t^m|x_{t-1},x_t), q_\phi (z_t^g | g_{t-1}, g_t)}  p_\theta (x_t, g_t | z_t, x_{t-1}, g_{t-1}) \\
& \ \ \ - D_{KL}(q_\phi (z_t^m | x_{t-1}, x_t) || p_\theta (z_t^m | x_{t-1})) \\
& \ \ \ - D_{KL}(q_\phi (z_t^g | g_{t-1}, g_t) || p_\theta (z_t^g | g_{t-1}))
\end{split}
\end{equation}
where, $D_{KL}(\cdot || \cdot)$ is the KL divergence between two distributions. In the above equation, we seek to minimize the KL divergence between the conditional priors and their corresponding posterior encoders (which are called motion/interaction prior in the main paper). For the expectation term on the right side of the equation, it measures the reconstruction quality of our shared decoder conditioned on the distributions of encoders. Therefore, our model is trained to minimize the loss function of 
\begin{equation}
\begin{split}
& \mathcal{L}_{recon_m} + \mathcal{L}_{recon_g} + \mathcal{L}_{KL_m} + \mathcal{L}_{KL_g} + \mathcal{L}_{consist} \\
& \mathcal{L}_{recon_m} = ||x_t - \hat{x}_t||^2, \ \mathcal{L}_{recon_g} = ||g_t - \hat{g}_t||^2 \\
\end{split}
\end{equation}
where $\hat{x}_t$, $\hat{g}_t$ are the reconstruction motion and interaction states from the decoder. $\mathcal{L}_{KL_m}$ and $\mathcal{L}_{KL_g}$ are the KL divergence loss terms for motion distribution and interaction distribution, respectively. We also add a consistency loss to promote consistency between the learned interaction state $\hat{g}_t$ and the human-ground interaction information, which is extracted through function $f(\cdot)$ from the predicted joints $\hat{x}_t$ and the ground plane $n$, $i.e.$,
\begin{equation}
\mathcal{L}_{consist} = ||\hat{g}_t - f(\hat{x}, n) ||^2
\end{equation}

\noindent\textbf{Inference.}
With the learned \bodymethodname model, there are two ways for inference. 
(a) Given the initial states $x_0$, $g_0$, sample latent variables $z_t^m$, $z_t^g$ from the conditional priors, and infer the remaining sequence of states $(x_1, g_1, \cdots, x_T, g_T)$ in an auto-regressive way by the shared decoder. (b) Given a sequence of noisy motion and interaction states, sample latent variables from the posterior encoders, and infer the motion and interaction states $x_1, g_1, \cdots, x_T, g_T$ by the shared decoder in a batch.

During training, we alternate (a) and (b) to improve the robustness of the CVAE model. During optimization, we adopt (b) to obtain the initial latent variable sequences from a sequence of motion and interaction states initialized by VPoser-t~\cite{vposer} and PARE~\cite{pare}.

\section{Experiment Details}\label{supp_sec:experiment}

\subsection{RGB Video Setting}

\noindent\textbf{Ablation study for initialization.}
In order to comprehensively validate the efficacy of \bodymethodname, we employ additional single-frame methods, namely FastMETRO~\cite{fastmetro} and CLIFF~\cite{cliff} as initialization techniques. As presented in Table~\ref{tab:unknown_ground_setting_results_different_inits}, the superior performance exhibited by \bodymethodname across various initialization methods validates its proficiency to handle the intricate human-ground relationship. It is noteworthy to mention that there exist contemporary cutting-edge single-frame methodologies that can be considered as potential initialization approaches, exemplified by \cite{nemo,niki,zhuo2023towards}.

\begin{table}
\begin{center}
\begin{tabular}{l|ccc}
\hline
Method & Cos ($\uparrow$) & Cos 1\% ($\uparrow$) & MPJPE* 1\% ($\downarrow$)\\
\hline
FM~\cite{fastmetro} + HuMoR~\cite{humor} & 0.99206 & 0.70771 & 417.8\\
FM~\cite{fastmetro} + \textbf{\bodymethodname} & \textbf{0.99965} & \textbf{0.92021} & \textbf{382.1} \\
CLIFF~\cite{cliff} + HuMoR~\cite{humor} & 0.99245 & 0.67262 & 345.5 \\
CLIFF~\cite{cliff} + \textbf{\bodymethodname} & \textbf{0.99965} & \textbf{0.84086} & \textbf{299.2} \\
\hline
\end{tabular}
\end{center}
\caption{Results on AIST++ dataset under the RGB video setting with initializaiton methods FastMETRO~\cite{fastmetro} (denoted ``FM'') and CLIFF~\cite{cliff}. Metrics are the same as Table~\ref{tab:unknown_ground_setting_results}.}
\label{tab:unknown_ground_setting_results_different_inits}
\end{table}

\section{Limitation and Future work}\label{supp_sec:limitation}

\begin{figure}
    \centering
    \includegraphics[width=\linewidth]{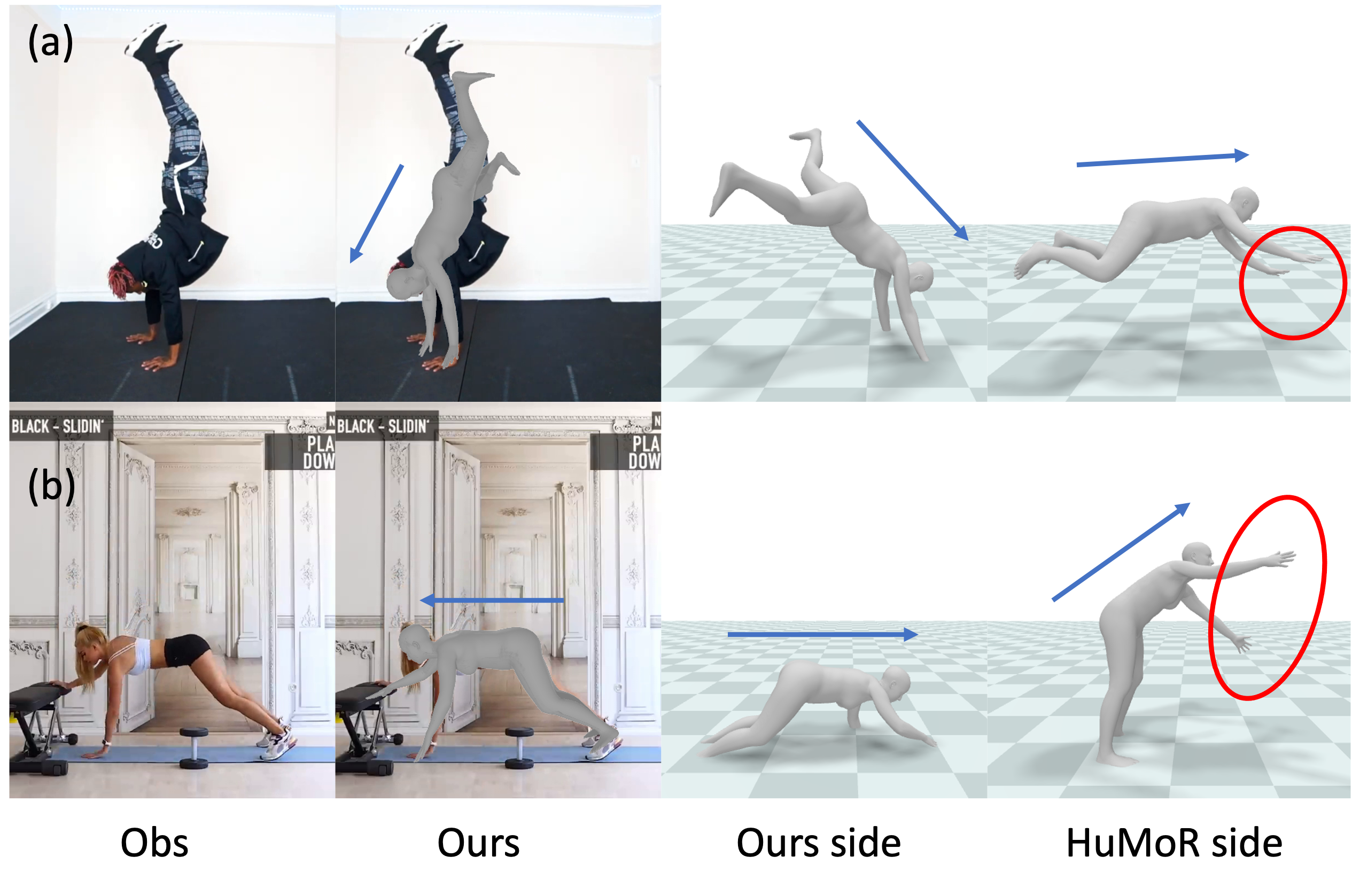}
    \caption{Failure cases. The \textcolor{HaveLockBlue}{direction of the body torso} and the \textcolor{red}{contact of HuMoR} are highlighted. Our method failed in the cases where the extreme angle and a second contact floor exist. But our prediction is still better than HuMoR's. ``side'' here means the side view in the world space.}
    \label{fig:failed_cases}
\end{figure}

Although our \bodymethodname model can yield superior performance in predicting physically plausible motion and reasonable ground planes in challenging cases, there are some limitations, such as hand inconsistency, limited contact joints. In some extreme cases as shown in Fig.~\ref{fig:failed_cases}(a), our model can make a reasonable inference on the ground plane but have a large error in positions due to the extreme angles and the high moving speed. Moreover, as shown in Fig.~\ref{fig:failed_cases}(b), for cases with more than one contact plane, our method is unable to separate the two contact planes. Nonetheless, our approach still outperforms the baseline method HuMoR in these challenging cases. 

In future work, it is promising to learn a stronger prior from large-scale training data (\textit{e.g.}, multiple contact surfaces, flexible contact joints, fine-grained hand motion) to further improve the performance. Furthermore, there is great potential for extension to enhance scene awareness and tackle occlusion observations, which are two long-standing challenges in the field of 3D human motion reconstruction. One possible avenue for improvement is to build upon the foundation of \bodymethodname by extending the continuous interaction representation to incorporate scene context, thereby enabling modeling of human-scene interaction. Additionally, we can incorporate a more robust initialization method specifically designed to handle occlusion cases. By delving deeper into these areas, we aim to advance the understanding and capabilities of 3D human motion reconstruction.

\section{Risks and potential misuse}\label{supp_sec:risk}

Since our techniques can generate realistic and diverse 3D human motion sequences from videos, there is a risk that such techniques could be potentially misused for fake video generation. We hope to raise the public’s awareness about the safe use of such technology. 

\end{document}